# DAPLSR: Data Augmentation Partial Least Squares Regression Model via Manifold Optimization

Haoran Chen*, Jiapeng Liu, Jiafan Wang and Wenjun Shi

School of Computer Science and Technology, Zhengzhou University of Light Industry, Zhengzhou 450000, China

## ABSTRACT

*Traditional Partial Least Squares Regression (PLSR) models frequently underperform when handling data characterized by uneven categories. To address the issue, this paper proposes a Data Augmentation Partial Least Squares Regression (DAPLSR) model via manifold optimization. The DAPLSR model introduces the Synthetic Minority Over-sampling Technique (SMOTE) to increase the number of samples and utilizes the Value Difference Metric (VDM) to select the nearest neighbor samples that closely resemble the original samples for generating synthetic samples. In solving the model, in order to obtain a more accurate numerical solution for PLSR, this paper proposes a manifold optimization method that uses the geometric properties of the constraint space to improve model degradation and optimization. Comprehensive experiments show that the proposed DAPLSR model achieves superior classification performance and outstanding evaluation metrics on various datasets, significantly outperforming existing methods.*

## 1. INTRODUCTION

Due to the swift advancement of computer technology and multimedia, pattern recognition is increasingly faced with complex, high-dimensional datasets, including images, text, and speech. The diversity and complexity of these data bring new challenges to the learning of machine learning models. However, during the acquisition process, it is common to encounter rich sample content but an insufficient number of samples.

Data augmentation is a widely adopted technique for the problem of small sample size, large number of categories and unevenness of categories. Data augmentation involves creating additional training samples by performing a sequence of changes and expansions to the original dataset. This process enhances the diversity and complexity of the data. These transformations include rotation, translation, scaling, flipping [1], and adding noise to simulate real-world data variations and uncertainties. Implementing data augmentation techniques enhances the model's capacity to generalize and robustness, hence mitigating the likelihood of overfitting, particularly in scenarios with scarce or imbalanced data.

As the field of the dimensionality of industrial data gradually increases, traditional modeling approaches may not fully utilize the potential information included in the data, especially when dealing with high-dimensional datasets. Data augmentation techniques have been shown to effectively improve prediction accuracy when combined with machine learning techniques such as Principal Component Analysis (PCA) [2], Support Vector Machines (SVMs) [3] [4], and Linear Discriminant Analysis (LDA) [5]. Inspired by these studies, this paper comb-ines data augmentation techniques with Partial Least Squares (PLS) methods [6] [7]. Partial Least Squares Regression (PLSR) is a classical machine learning method used to model linear relationships between variables [8]. It is widely applied in modeling original data with small samples, high-dimensional features, while data augmentation can compensate for the insufficient sample size, as well as the imbalance of categories in the original data [9] [10] [11].

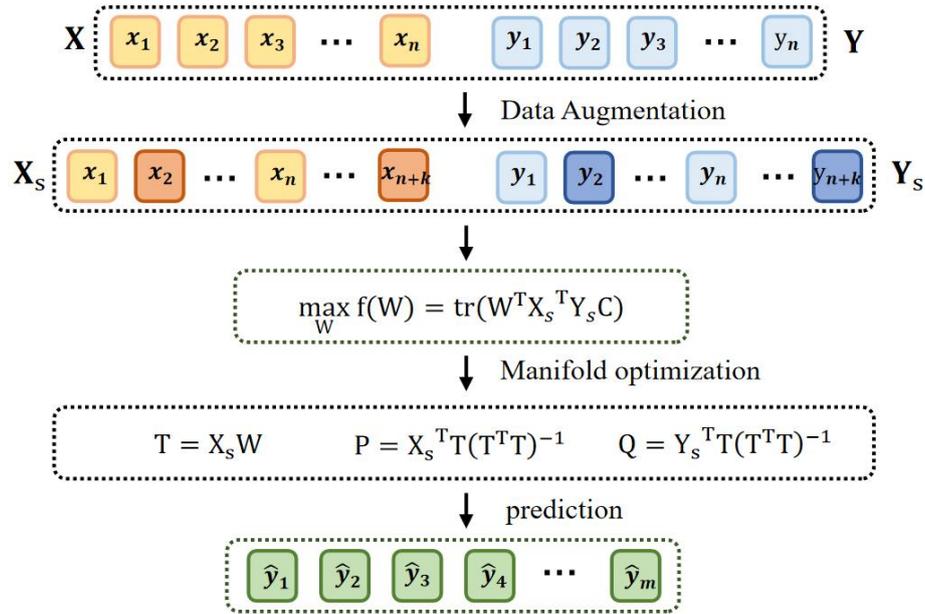

Figure 1.   Diagram of the DAPLS modeling framework

This study uses the Synthetic Minority Over-sampling Technique (SMOTE) [12] to address the problem of insufficient sample size and category imbalance. The amount of intrinsic irrelevant attributes in the data increases with dataset complexity when applying SMOTE to original data samples. This can adversely affect the accuracy of the categorization method [13]. It is crucial to automatically detect and prevent the impact of these factors on the categorization process. Therefore, the Value Difference Metric (VDM) [14] is also introduced to identify and evaluate which features are more critical for the effectiveness of the oversampling method. VDM, as a measure of sample similarity, aids in selecting neighboring samples with high similarity to the original samples, ensuring that the synthetic samples generated maintain the feature distribution and attributes of the original samples while minimizing overlap between synthetic and original samples [15]. Following the augmentation of the original data, the PLSR model is employed for mathematical modeling. During the model solving process, the objective function is solved using manifold optimization to achieve a more accurate numerical solution, which in turn enhances the model's classification and recognition performance.

By integrating data augmentation techniques and partial least squares methods, the new model can better learn data features and improve the recognition capability for minority class samples. The new model not only demonstrates superior numerical solutions but also has better interpretability and efficiency, providing novel ideas and approaches for handling high-dimensional data and uncovering latent information. Figure 1 illustrates the framework of the DAPLS model.

The main contributions of this paper are as follows:

This paper combines data augmentation techniques with the PLSR model to improve the model's learning ability and classification performance on small samples and unbalanced category data. The Synthetic Minority Over-sampling Technique is used to generate synthetic samples and is combined with the Value Difference Metric to select similar neighboring samples, thereby improving the quality of the synthetic samples and effectively solving the problems of insufficient sample size and class imbalance.

This paper proposes manifold optimization, which is an unconstrained optimization method in a restricted search space, to obtain more accurate numerical solutions of PLSR.

The DAPLSR model is compared with five existing models on a public dataset and analyzed in terms of the effect of the number of different components retained on the performance of the model. The experimental results show that the proposed model has a significant advantage in classification performance, especially when dealing with unbalanced datasets.

## 2. RELATED WORKS

### 2.1. PLSR

Suppose the training dataset consists of an input matrix $\mathbf{X} = \left[ \mathbf{x}_1, \mathbf{x}_2, \ldots, \mathbf{x}_{N+p} \right] \in \mathbb{R}^{m \times (N+p)}$. and an output matrix $\mathbf{Y} = \left[ \mathbf{y}_1, \mathbf{y}_2, \ldots, \mathbf{y}_n \right] \in \mathbb{R}^{n \times q}$, where $m$ denotes the number of dimensions and $n$ denotes the number of samples.

To initially normalize the data, the data samples are removed by their average. $\mathbf{X}$ and $\mathbf{Y}$ are projected into a low-dimensional subspace by the PLSR model, which is defined by the latent variables $\mathbf{T}$ and $\mathbf{U}$ of $\mathbf{X}$ and $\mathbf{Y}$ [16]. The PLSR model is then used as the basis for the normalization of the data. Let $\mathbf{t}_1$ and $\mathbf{u}_1$ represent the initial pair of latent variables for $\mathbf{X}$ and $\mathbf{Y}$, respectully. These variables can be written sas the following linear combinations:

$$\begin{aligned} \mathbf{t}_1 &= \mathbf{X}\mathbf{w}_1 \\ \mathbf{u}_1 &= \mathbf{Y}\mathbf{c}_1 \end{aligned} \tag{1}$$

the objective of PLSR is to ensure that $\mathbf{t}_1$ and $\mathbf{u}_1$ include the maximum amount of relevant information from the raw data, hence maximizing their relevance. As a result, the objective function can be calculated by first maximizing the covariance between $\mathbf{t}_1$ and $\mathbf{u}_1$ in PLS:

$$\begin{aligned} \max & \left\langle \mathbf{X}\mathbf{w}_1, \mathbf{Y}\mathbf{c}_1 \right\rangle \\ \textbf{s.t. } & \mathbf{w}_1^\mathrm{T}\mathbf{w}_1 = 1, \mathbf{c}_1^\mathrm{T}\mathbf{c}_1 = 1 \end{aligned} \tag{2}$$

where are the initial weight vectors, $\mathbf{w}_1$ and $\mathbf{c}_1$. $\mathbf{w}_1$ and $\mathbf{c}_1$ can be solved by solving for the eigenvectors corresponding to the largest eigenvalues of matrices $\mathbf{X}^\mathrm{T}\mathbf{Y}\mathbf{Y}^\mathrm{T}\mathbf{X}$ and $\mathbf{Y}^\mathrm{T}\mathbf{X}\mathbf{X}^\mathrm{T}\mathbf{Y}$, respectively, which in turn can be solved for $\mathbf{t}_1$ and $\mathbf{u}_1$ by using Eq.(1). This model allows the PLSR to be expressed as a pair of matrices, X and Y:

$$\mathbf{X} = \mathbf{t}_1 \mathbf{p}_1^T + \mathbf{E}_1$$
$$\mathbf{Y} = \mathbf{u}_1 \mathbf{q}_1^T + \mathbf{F}_1$$

(3)

where the load vectors $\mathbf{p}_1 = \mathbf{X}^T \mathbf{t}_1 \left( \mathbf{t}_1^T \mathbf{t}_1 \right)^{-1}$ and $\mathbf{q}_1 = \mathbf{Y}^T \mathbf{t}_1 \left( \mathbf{t}_1^T \mathbf{t}_1 \right)^{-1}$ are calculated as regression coefficients of $\mathbf{X}$ on $\mathbf{t}$ and $\mathbf{Y}$ on $\mathbf{u}$, respectively [17]. The method can be derived using the least squares approach. There are two residual matrices: $\mathbf{E}$ and $\mathbf{F}$. PLSR utilizes an iterative approach to the computation of latent variables [18], where the latent variables, or principal components, are repeatedly included in the model.

$$\mathbf{X}^* = \mathbf{X} - \mathbf{t}_1 \mathbf{p}_1^T$$
$$\mathbf{Y}^* = \mathbf{Y} - \mathbf{u}_1 \mathbf{q}_1^T$$

(4)

Following several iterations of the previously described steps, PLSR ultimately produced the following regression model:

$$\mathbf{X} = \mathbf{T}\mathbf{P}^T + \mathbf{E}$$
$$\mathbf{Y} = \mathbf{U}\mathbf{Q}^T + \mathbf{F}$$
$$\mathbf{Y} = \mathbf{X}\mathbf{R} + \mathbf{F}$$

(5)

where $\mathbf{T}$ and $\mathbf{U}$ are the score $\mathbf{X}$ and $\mathbf{Y}$, matrix $\mathbf{E}$ and matrix $\mathbf{F}$ are the residual matrices and $\mathbf{R} = \mathbf{W}\mathbf{Q}^T$ can be thought of as a regression coefficient matrix.

## 2.2. Data Augmentation

In dealing with the problems of small sample size, large number of categories, and imbalance of categories, researchers and data scientists have used a variety of methods to improve the recognition performance of the models. One common approach is to expand the training dataset through data augmentation techniques, such as rotating, flipping, scaling, and other transformations to enhance the variety of samples.

This approach can successfully mitigate the issue of inadequate samples and enhance the model's generalization capability. When the amount of data is much smaller than the model's demand for unknown sample space, the model's generalization effect may be affected. To solve this problem, data augmentation methods become an effective means. As in the case of Ensemble Learning, the prediction results of multiple base classifiers are combined to improve the overall classification performance. Common integration learning methods include Random Forest, Gradient Boosting Trees, and so on. These methods can effectively deal with category imbalance and small sample size to improve the robustness and precision of the model. Data augmentation methods simulate the original data to generate virtual data and labels. These are then added to the training dataset, thus increasing its size.

Another common approach to data augmentation is to use training methods based on sample weights, such as oversampling and undersampling. Oversampling increases the proportion of minority class samples by replicating minority class samples or generating synthetic samples, while undersampling balances the number of samples in different classes by removing majority class samples. These methods can help the model learn better about the distribution of the data and improve the ability to recognize samples from minority classes.

SMOTE is an oversampling technique used to solve the problem of category imbalance [19]. In machine learning, category imbalance refers to a large gap in the number of samples from different categories in a dataset, which may lead to over-concentration of the model's learning on the majority category while ignoring the minority category [20]. The objective of SMOTE is to improve the model's prediction performance for the minority category by generating fresh samples of the minority category, thereby equalizing the distribution of samples across different categories.

SMOTE does this by selecting a sample from a small number of categories as a benchmark sample. For the benchmark sample, its nearest neighbor is selected based on some distance metric (usually using Euclidean distance). For the selected baseline and nearest neighbor samples, a point is randomly selected on the connecting line between them as the newly synthesized sample. The eigenvalues of this new sample are obtained from the weighted average of the eigenvalues of the baseline and nearest neighbor samples, and the category label is the minority category. Repeat the resampling many times until a specified oversampling multiplier is reached. In this way, SMOTE allows the proportion of minority categories in the dataset to increase by generating new synthetic samples. This helps the model to learn the features of the minority categories better and improves the performance on the unbalanced dataset.

## 3. DATA AUGMENTATION PARTIAL LEAST SQUARES REGRESSION

The method described employs the SMOTE data augmentation methodology to generate more samples, thereby equalizing the quantity of samples across different categories in the dataset, and uses the VDM to calculate distances between attributes in order to allow the new data samples to maintain the feature distributions and variability of the original data. The sampled data $\mathbf{X}_s$ and $\mathbf{Y}_s$ are used as new data samples, the covariance matrices of $\mathbf{X}_s$ and $\mathbf{Y}_s$ are computed using the partial least square regression model. The PLS components are calculated by maximizing the covariance matrices of $\mathbf{X}_s$ and $\mathbf{Y}_s$. In seeking to maximize the covariance matrices of $\mathbf{X}_s$ and $\mathbf{Y}_s$, the local geometrical structure of the manifolds is exploited, and the use of manifold optimization techniques enables the objective function to obtain the maximum value on the manifolds, thereby enhancing the model's performance.

$\mathbf{X} = [\mathbf{x}_1, \mathbf{x}_2, ..., \mathbf{x}_n] \in \mathbb{R}^{n \times d}$, where $n$ and $d$ stand for the number of samples and sample features, is the representation of the sample input data matrix. The labeling matrix is represented as $\mathbf{Y} = [\mathbf{y}_1, \mathbf{y}_2, ..., \mathbf{y}_n] \in \mathbb{R}^{n \times q}$, where $q$ denotes the quantity of sample categories, $\mathbf{x}_i$ denotes the first $i$ sample, and for each minority category sample $\mathbf{x}_i$, k similar samples centered on $\mathbf{x}_i$ are selected, and the one sample as the reference sample. For each feature dimension $j$, a value difference metric matrix $\mathbf{D}$ is used to determine the values of the synthetic samples based on the difference between the reference sample and the nearest neighbor sample, and then the dummy data are generated by interpolation operation.

$$\mathbf{x}_{si} = \mathbf{x}_i + \varsigma \times \{\mathbf{x}_i - s([\mathbf{x}_1, \mathbf{x}_2, ..., \mathbf{x}_n])\} \tag{6}$$

where $\varsigma$ denotes a random number between $(0,1)$, and $s(\cdot)$ refers to random sampling in the data sample. The newly generated samples and the original samples from the new data, so the data samples can be denoted as $\mathbf{X}_s = \left[\mathbf{x}_1, \mathbf{x}_2, ..., \mathbf{x}_{si}, ..., \mathbf{x}_{n+m_i}\right] \in \mathbb{R}^{(n+m_i) \times d}$ after random sampling, and the array of new sample labels generated using the k-nearest neighbor is $\mathbf{Y}_s = \left[ \mathbf{y}_1, \mathbf{y}_2, ...., \mathbf{y}_{si}, ..., \mathbf{y}_{n+m_l} \right] \in \mathbb{R}^{(n+m_l) \times q}$, where $m_1$ denotes the number of freshly created samples.

During the process of generating new data samples as described above, the Value Difference Metric Matrix D uses the Value Difference Metric, which helps SMOTE determine how to synthesize appropriate samples that maintain the characteristic distribution and variability of the original data. The distance between two feature values is ascertained by contrasting the class conditional probability distributions of each feature's values $i_a$ and $j_a$. This makes it different from other distance measures and provides an alternate way for calculating the distance between two symbolic values.

$$\mathbf{vdm}(i, j) = \sum_{a=0}^{d} \delta(i_a, j_a) \cdot \omega(i_a)$$
$$\delta(i_a, j_a) = \sum_{c \in C} | \mathbf{P}(c \mid i_a) - \mathbf{P}(c \mid j_a) |^2 \qquad (7)$$
$$\omega(i_a) = \left[ \sum_{c \in C} | \mathbf{P}(c \mid i_a) |^2 \right]$$

$\delta(i_a, j_a)$ indicates whether the values of feature $i$ and feature $j$ at position $a$ are identical. If identical, $\delta(i_a, j_a) = 0$; if different, $\delta(i_a, j_a) = 1$. C refers to the set of all class labels, $\mathbf{P}(c \mid i_a)$ refers to the class conditional probability of $i_a$, and $\omega(i_a)$ represents the weight of the feature $i$ at position $a$, controlling the influence of each attribute distance in determining the final nearest neighbor. Since the values of $\delta(i_a, j_a)$ the range of values will vary from 0 and 1, the weights can be used to limit that range, i.e., the range of the $\delta(i_a, j_a) \cdot \omega(i_a)$ will vary from 0 and $\omega(i_a)$. The $\mathbf{vdm}(i, j)$ value is more significantly impacted by larger attribute distances than by smaller attribute distances. Therefore, when using smaller weights (i.e., when the values in the tested instance are not associated), the outcome of attribute distances tends to be tiny and does not significantly affect the selection of nearest neighbors. VDM leverages the weight of values to ascertain the capacity of specific attribute values to distinguish between class labels. Summing the difference metrics across all positions yields the feature $i$ and feature $j$ value difference metrics between them.

After obtaining the value difference metric $\mathbf{vdm}(i, j)$ for each pair of features, all $\mathbf{vdm}(i, j)$ values are compiled into a matrix $\mathbf{D}$, where $\mathbf{D}$ is a symmetric matrix with $\mathbf{D}_{ij} = \mathbf{D}_{ji}$. This matrix is utilized in SMOTE to aid in generating more representative and diverse synthetic samples.

PLSR maximizes the covariance between the transformed $\mathbf{X}$ and $\mathbf{Y}$ by finding a linear transformation of $\mathbf{X}$ and $\mathbf{Y}$. The $\mathbf{T}$ is obtained by principal component analysis (PCA) of $\mathbf{X}$, which can be expressed as $\mathbf{T} = \mathbf{XW}$, where $\mathbf{W}$ is the projection matrix of $\mathbf{X}$ in the component space. Meanwhile, one of the most important modeling advantages of PLSR is to seek the optimal projection matrix to make the relationship between $\mathbf{X}$ and $\mathbf{Y}$ the closest by considering the projections of the independent variable and the dependent variable simultaneously. Then, according to the previous description, $\mathbf{W}$ can be identified as the projection matrix of $\mathbf{X}$ in the component space, and $\mathbf{C}$ is the projection matrix of $\mathbf{Y}$ in the component space, denoted as $\mathbf{U} = \mathbf{YC}$, which serves as the SIMPLSR model equivalent [21], and the following is the expression for the matching objective function:

$$\max_{\mathbf{W},\mathbf{U}} \ \text{tr}(\mathbf{W}^{\mathrm{T}}\mathbf{X}^{\mathrm{T}}\mathbf{Y}\mathbf{C})$$
$$\text{s.t. } \mathbf{W}^{\mathrm{T}}\mathbf{W} = \mathbf{C}^{\mathrm{T}}\mathbf{C} = \mathbf{I} \tag{8}$$

Using the data-enhanced sample as the new input data for PLS, the new objective function can be expressed as:

$$\max_{\mathbf{W}} \ \text{tr}(\mathbf{W}^{\mathrm{T}}\mathbf{X}_s^{\mathrm{T}}\mathbf{Y}_s\mathbf{C})$$
$$\text{s.t. } \mathbf{W}^{\mathrm{T}}\mathbf{W} = \mathbf{C}^{\mathrm{T}}\mathbf{C} = \mathbf{I} \tag{9}$$

This study utilizes the manifold optimization approach to solve the optimal solution of the objective function. In Eq.(9), the feasible regions for the projection matrices $\mathbf{W}$ and $\mathbf{C}$ can be regarded as a product manifold composed of an Oblique manifold and a generalized Stiefel manifold [22]. Consequently, the new objective function can be expressed in the following form:

$$\max_{\mathbf{W},\mathbf{C}} \ \text{tr}(\mathbf{W}^{\mathrm{T}}\mathbf{X}_s^{\mathrm{T}}\mathbf{Y}_s\mathbf{C})$$
$$\text{s.t. } \mathbf{T}^{\mathrm{T}}\mathbf{T} = \mathbf{W}^{\mathrm{T}}\mathbf{X}^{\mathrm{T}}\mathbf{X}\mathbf{W} = \mathbf{I} \tag{10}$$
$$\text{diag}(\mathbf{C}^{\mathrm{T}}\mathbf{C}) = \mathbf{I}$$

First, to calculate the objective function's gradient in Euclidean space with respect to the projection matrix $\mathbf{W}$ of $\mathbf{X}$:

$$\text{grad}_E f(\mathbf{W}) = \mathbf{X}_s^{\mathrm{T}}\mathbf{Y}_s\mathbf{C} \tag{11}$$

Next, the gradient on the generalized Stiefel manifold is calculated:

$$\text{grad}_R f(\mathbf{W}) = \text{grad}_E f(\mathbf{W}) - \mathbf{W}\text{symm}(\mathbf{W}^{\mathrm{T}}\mathbf{B}\text{grad}_E f(\mathbf{W})) \tag{12}$$

The above equation is based on the gradient of the objective function in the Euclidean space, and the gradient projected to the tangent space by the projection operator is the gradient on the manifold. The above equation is based on the gradient of the objective function in the Euclidean space, and the gradient projected to the tangent space by the projection operator is the gradient on the manifold. Where the expression of the projection operator is $\mathcal{P}_{[\mathbf{W}]}(\mathbf{W}) = \mathbf{W} - \mathbf{Z}\text{symm}(\mathbf{Z}^{\mathrm{T}}\mathbf{B}\mathbf{W})$, where $\mathbf{B}$ is a symmetric positive definite matrix [23], and $\text{symm}(\mathbf{A}) = (\mathbf{A} + \mathbf{A}^{\mathrm{T}})/2$.

In the process of solving the gradient descent on the manifold, the newly discovered points may not lie on the manifold. In such cases, the retraction operator $\mathcal{R}_x$ is needed to project the points onto the manifold pointing from the tangent space [24], and the optimal projection matrix $\mathbf{W}$ is obtained step by step according to the set step size.

The optimized projection matrix $\mathbf{W}$ is brought into the gradient formulation of the objective function with respect to the projection matrix $\mathbf{Y}$ of $\mathbf{C}$ in Euclidean space:

$$\text{grad}_E f(\mathbf{C}) = \mathbf{Y}_s^{\mathrm{T}}\mathbf{X}_s\mathbf{W} \tag{13}$$

Calculate the gradient over the Oblique manifold:

$$\text{grad}_R f(\mathbf{C}) = \text{grad}_E f(\mathbf{C}) - \mathbf{C} \, \text{diag}(\mathbf{C}^T \text{grad}_E f(\mathbf{C})) \qquad (14)$$

Similar to the optimization process for the projection matrix $\mathbf{W}$, the expression of the projection operator for the Oblique manifold tangent space is $\mathcal{P}_{[\mathbf{C}]}(\mathbf{Z}) = \mathbf{Z} - \mathbf{C} \text{diag}(\mathbf{C}^T \mathbf{Z})$, The retraction operator $\mathcal{R}_x$ is also needed to project the points onto the manifold pointing from the tangent space after a new point is found on the Oblique manifold, and the solution is iteratively solved until a minima is obtained, resulting in the optimal projection matrix $\mathbf{C}$.

Using the Alternating Direction Method of Multipliers (ADMM) [25]. The optimization of the projection matrices $\mathbf{W}$ and $\mathbf{C}$ is executed in the objective function. The optimal projection matrices $\mathbf{W}$ and $\mathbf{C}$ are computed, then PLSR is conducted using the obtained matrices $\mathbf{W}$. With the use of manifold optimization, we suggest a data augmentation partial least squares technique in Algorithm 1.

---

## Algorithm 1 DAPLSR

1: Initialize: $\mathbf{X} \in \mathbb{R}^{n \times m}$, $\mathbf{Y} \in \mathbb{R}^{n \times q}$, number of iterations $\mathbf{N}$, the gradient standardized tolerance $\epsilon_1$ and the step error $\epsilon_2$. Use the VDM to compute the value difference metric matrix $\mathbf{D}$ for all features in the feature space

2: Synthesizing new samples: For each category $c$ in the dataset, calculate its sample size $\mathbf{n}_c$, and determine the set of minority class samples $\mathbf{X}_{min}$ and the corresponding set of labels $Y_{min}$. For each minority class sample $\mathbf{x}_i \in \mathbf{X}_{min}$, find its k nearest neighbors. Generate synthetic samples based on its nearest neighbor samples and value difference metric matrix $\mathbf{D}$. Generate new synthetic sample labels based on the minority sample labels

3: Add the synthesized samples to the dataset to get new data samples $\mathbf{X}_s$ and data labels $\mathbf{Y}_s$

4: **for** $j = 1: N$ **do**

5:     **for** $l = 1: N$ **do**

6:         Calculate the objective function's gradient in Euclidean space with respect to the projection matrix $\mathbf{W}$ of $\mathbf{X}$: $\text{grad}_E f(\mathbf{W}_l) = 2\mathbf{X}_s^T \mathbf{Y}_s \mathbf{C}_l$

7:         Compute the gradient of the objective function on the generalized Stiefel manifold, where $\mathcal{P}_{[\mathbf{W}]}$ is the projection matrix in tangent space: $\text{grad}_R f(\mathbf{W}_l) = \mathcal{P}_{[\mathbf{W}]} \text{grad}_E f(\mathbf{W}_l)$

8:         Calculate the conjugate direction: $\zeta_l = -\text{grad}_R f(\mathbf{W}_l) + \beta_l \mathbf{T}_{\mathbf{W}_{l-1} \to \mathbf{W}_l}(\zeta_{l-1})$

9:         Setting the step size $\alpha_l$: $f(\mathbf{R}_{\mathbf{W}_l}(\alpha_l \zeta_l)) \geq f(\mathbf{W}_l) + \alpha_l \text{tr}((\text{grad}_R f(\mathbf{W}_l))^T \zeta_l)$

10:        The loop is terminated when the condition is satisfied by $\mathbf{W}_l = \mathbf{R}_{\mathbf{W}_l}(\alpha_l \zeta_l)$, and $\left\| (\text{grad}_R f(\mathbf{W}_{l+1}) \right\|_F \leq \epsilon_1$, $\alpha_1 \leq \epsilon_2$ and $l \geq N_l$

11:     **end for**

12:     Exports $\mathbf{W} = \mathbf{W}_l$

13:     **for** $k = 1: N_2$ **do**

14:         Calculate the objective function's gradient in Euclidean space with respect to the projection matrix $\mathbf{C}$ of $\mathbf{Y}$: $\text{grad}_E f(\mathbf{C}_k) = \mathbf{Y}_s^T \mathbf{X}_s \mathbf{W}$

15:       Compute the gradient of the objective function on the generalized Stiefel manifold:

$$\text{grad}_{\mathbb{R}} f(\mathbf{C}_k) = \mathcal{P}_{[C]} \text{grad}_{\mathbb{E}} f(C_k)$$

16:       Calculate the conjugate direction: $\zeta_k = -\text{grad}_{\mathbf{R}} f(\mathbf{C}_k) + \beta_l \mathbf{T}_{\mathbf{C}_{k-1} \to \mathbf{C}_k} (\zeta_{k-1})$

17:       Setting the step size $a_k$ : $f(\mathbf{R}_{k_l}(\alpha_k \zeta_k) \ge f(\mathbf{C}_k) + \alpha_k \text{tr}((\text{grad}_{\mathbf{R}} f(\mathbf{C}_k))^{\mathrm{T}} \zeta_k)$

18:       The loop is terminated when the condition is satisfied by $\mathbf{C}_k = \mathbf{R}_{\mathbf{C}_k}(\alpha_k \zeta_k)$ and

                 $\|(\text{grad}_{\mathbf{R}} f(\mathbf{C}_{k+1})\|_{\mathbf{F}} \le \epsilon_1, \ \alpha_k \le \epsilon_2$ and $k \ge N_2$

19:   **end for**

20:    Exports $\mathbf{C} = \mathbf{C}_k$

21: **end for**

22: Calculate: $\mathbf{T} = \mathbf{X}_s \mathbf{W}$          $\mathbf{P} = \mathbf{X}_s{}^{\mathrm{T}} \mathbf{T}(\mathbf{T}^{\mathrm{T}} \mathbf{T})^{-1}$          $\mathbf{Q} = \mathbf{Y}_s{}^{\mathrm{T}} \mathbf{T}(\mathbf{T}^{\mathrm{T}} \mathbf{T})^{-1}$

**Output**: $\mathbf{Y} = \mathbf{X}_s \mathbf{W} \mathbf{Q}$

---

# 4. EXPERIMENTS

This study conducts experiments on the EYaleB dataset, COIL-20 object dataset, USPS handwritten digit dataset, and Brodatz texture dataset. Using classification accuracy, G-mean, F-measure, Precision, and Recall as metrics, the proposed method's performance is thoroughly assessed. The experimental procedure selects five models—NIPALS [26], PolyPLSR [27], SIMPLSR, SPLPLSR [28], and PLSRGGR [29]—as benchmark models for assessing the classification performance of the proposed model on various public datasets.

The MATLAB 2019b software platform is utilized for all experiments, which are executed on a high-performance computing environment that is outfitted with an Intel E5 2620 CPU and up to 128 GB RAM. This robust computational assistance guarantees the precision and dependability of the experimental outcomes. To ensure fairness, the number of components for different models is kept consistent across each dataset for rigorous experimental comparison. Evaluation metrics are compared only when all methods demonstrate satisfactory performance. Each experiment is repeated five times, and the average of these five results is taken as the final evaluation basis to ensure stability and reliability.

## 4.1. Evaluation indicators

In classifier evaluation, the confusion matrix is a commonly used tool that provides a detailed analysis of classifier performance.

The confusion matrix includes four key metrics: TN (True Negatives) refers to the number of correctly categorized negative samples, FP (False Positives) represents the number of negative samples incorrectly categorized as positive, FN (False Negatives) denotes the number of positive samples that are incorrectly classified as negative. TP (True Positives) indicates the number of samples that are correctly classified.

In multicategorization problems, these metrics are interpreted slightly differently. TP denotes the count of properly predicted minority samples, TN denotes the count of erroneously forecasted majority samples, FP denotes the count of mistakenly predicted majority samples, and FN denotes the count of incorrectly predicted minority samples. Metrics such as the confusion matrix and accuracy can provide a thorough evaluation of the classifier's performance, allowing us to get a more profound comprehension of the model's predictive effectiveness. The accuracy rate can be represented using the following equation:

$$\text{Accuracy} = \frac{\text{TP} + \text{TN}}{\text{TP} + \text{FP} + \text{TN} + \text{FN}} \qquad (15)$$

Mean (Geometric Mean) is an evaluation metric more applicable to unbalanced datasets. It combines the performance of the model in predicting the majority sample class as well as the minority sample class, and evaluates the overall performance of the model through the geometric mean between these two. The formula for G-Mean is as follows:

$$\text{G-Mean} = \sqrt{(\text{Sensitivity} \times \text{Specificity})} \qquad (16)$$

where Sensitivity (also known as Recall) is a measure of the model's capacity to recognize a small number of samples, and Specificity (also known as true-negative) measures the model's ability to recognize a large number of samples. These indicators are computed using the following formulas:

$$\text{Sensitivity} = \frac{\text{TP}}{\text{TP} + \text{FN}}$$
$$\text{Specificity} = \frac{\text{TN}}{\text{FP} + \text{TN}} \qquad (17)$$

Precision is a quantitative measure utilized to evaluate the efficiency of a classification model. It specifically measures the proportion of predicted positive examples that are actually positive cases. Precision has a value between 0 and 1, and the closer it is to 1, the better the model predicts positive cases. Precision is calculated as follows:

$$\text{Precision} = \frac{\text{TP}}{\text{TP} + \text{FP}} \qquad (18)$$

Recall values are calculated in a manner comparable to sensitivity, with a range of 0 to 1. The model performs better at predicting positive instances when the value is closer to 1.

$$\text{Recall} = \frac{\text{TP}}{\text{TP} + \text{FN}} \qquad (19)$$

The F-measure, which is a weighted summed average of Precision and Recall, is used to analyze the classification model's performance and aids in a more thorough evaluation of the model's performance in various application scenarios. The calculation is performed in the following manner:

$$\text{F-measure} = 2 \times (\frac{\text{Precision} \times \text{Recall}}{\text{Precision} + \text{Recall}}) \qquad (20)$$

The F-measure is a numerical value that runs from 0 to 1. Higher values indicate that the model achieves a better balance between Precision and Recall. Because it is a harmonic mean, the F-measure will be more sensitive to the smaller of the two values, so it provides a comprehensive metric when the model needs to balance Precision and Recall.

## 4.2. Introduction to the dataset

### 4.2.1. Face

For the comparison experiments, the Extended Yale Face Database B (EYaleB) dataset [30] is employed for the face classification task. During the experiment, 75% of the photos are allocated as training samples, while the remaining 25% are designated as testing samples. Samples from the dataset's 38 categories or persons are randomly chosen for training, with the unselected samples being used for testing, to mimic class imbalance in real-world applications.

To ensure consistency in processing, all images are vectorized and shrunk to $32 \times 32$ pixels before to the experiment. This results in 1024 feature components per sample.

### 4.2.2. Object

To verify the model's effectiveness in object image classification tasks, the dataset utilized is the Columbia Object Image Library (COIL-20) [31]. The dataset contains 20 object categories, each with 72 corresponding images. In the experiment, a varying number of images from each category are randomly selected as training and testing samples, ensuring a constant total sample number.

### 4.2.3. Handwritten digits

In addition to object and face image datasets, this study extends the experiment by selecting the USPS dataset [32] for the task of classifying handwritten digits. For the USPS dataset, a rich experimental sample set is constructed by randomly selecting 300 images from the original image samples in each category 0-9, with the number of samples in each category selected at random. Keeping in line with the previous processing, the number of features of each image is transformed into a 256-dimensional vector in the experiment for model training and recognition. To ensure a thorough assessment of the model's performance, the dataset is randomly split into two portions: one half is allocated for training, and the other half is reserved for testing.

### 4.2.4. Texture

The experiment assesses the efficacy of the model by subjecting it to a texture classification task, specifically employing the Brodatz dataset [33]. The Brodatz texture collection has 999 texture photos, which are categorized into 111 distinct categories, each containing 9 images. The images depict a diverse range of all-natural substances, including wood, stone, grass, and bark, as well as man-made substances like paper, gauze, and silk screen. During the experimental preparation phase, a certain number of images are randomly selected from each of the 111 categories of original image samples as training samples, with the remaining images serving as testing samples. The ratio of training to testing samples is set at 6:3 to simulate the distribution that may occur in real-world applications. Similarly, each sample image is converted into an 1180-dimensional vector to enable effective feature extraction and classification by the model.

## 4.3. Analysis of experimental results

Table 1 presents the classification outcomes of the DAPLSR model and its comparative methods on the EYaleB dataset, retaining different numbers of features. The bold numbers indicate the best results in the comparative experiments. On the EYaleB dataset, the suggested solution regularly performs better than alternative approaches in terms of classification error rates. Strong performance of the DAPLSR model is found to be maintained when the number of retained characteristics rises. Notably, when retaining 34 components, the proposed model's classification error rate is reduced by approximately 0.51% compared to the model without data augmentation, demonstrating the efficacy of incorporating data augmentation techniques in enhancing classification efficiency.

Table 1.   EYaleB Dataset Classification Error Rates

| c | 20 | 22 | 24 | 26 | 28 | 30 | 32 | 34 |
|---|----|----|----|----|----|----|----|----|
| NIPALS | 0.4780 | 0.4475 | 0.3729 | 0.3492 | 0.3644 | 0.3339 | 0.3288 | 0.3102 |
| PolyPLSR | 0.4508 | 0.4068 | 0.3763 | 0.3441 | 0.3305 | 0.2949 | 0.2898 | 0.2864 |
| SIMPLSR | 0.3797 | 0.3475 | 0.339 | 0.3119 | 0.2898 | 0.2746 | 0.2627 | 0.2559 |
| SPLPLSR | 0.3169 | 0.2814 | 0.2593 | 0.2373 | 0.2339 | 0.1966 | 0.2000 | 0.1814 |
| PLSRGGR | 0.1542 | 0.1220 | 0.1186 | 0.0915 | 0.0729 | 0.0492 | 0.0441 | 0.0322 |
| DAPLSR | **0.1441** | **0.0932** | **0.0712** | **0.0712** | **0.0407** | **0.0356** | **0.0407** | **0.0271** |

Table 2 illustrates the accuracy, G-mean, Precision, Recall, and F-measure values for various approaches on the EYaleB dataset. The outcomes demonstrate that the model described in this study maintains high values for all other metrics while maintaining high classification accuracy on the EYaleB dataset classification task, and all other metrics outperform the benchmark methods. These results underscore the effectiveness and feasibility of integrating data augmentation techniques, highlighting the stability and advantage of the DAPLSR model in handling face data classification tasks.

Table 2.    EYaleB Dataset Performance Metrics

| Method | Accuracy | G-mean | Precision | Recall | F-measure |
|--------|----------|--------|-----------|--------|-----------|
| NIPALS | 0.6661 | 0.8042 | 0.7959 | 0.6718 | 0.6886 |
| PolyPLSR | 0.7051 | 0.8307 | 0.8200 | 0.7107 | 0.7254 |
| SIMPLSR | 0.8153 | 0.8960 | 0.8455 | 0.8182 | 0.8117 |
| SPLPLSR | 0.7983 | 0.8856 | 0.8275 | 0.8010 | 0.7949 |
| PLSRGGR | 0.9441 | 0.9639 | 0.9565 | 0.9456 | 0.9365 |
| **DAPLSR** | **0.9627** | **0.9806** | **0.9657** | **0.9637** | **0.9626** |

Table 3 shows that on the COIL-20 dataset, the proposed DAPLSR model generally yields better recognition and classification results for objects across different numbers of retained components compared to other methods. Specifically, when retaining 15 components, the classification error rate decreases by 2.36% compared to the second-best model. With 21 components, while the second-best model achieves its lowest classification error rate of 2.92%, the proposed model further reduces the error rate by 0.14%. These findings indicate that applying data augmentation to the PLSR model significantly reduces classification error rates and enhances classification performance.

Table 3.    COIL-20 Dataset Classification Error Rates

| c | 13 | 15 | 17 | 19 | 21 | 23 |
|---|----|----|----|----|----|----|
| NIPALS | 0.2111 | 0.2319 | 0.1792 | 0.1861 | 0.1542 | 0.1347 |
| PolyPLSR | 0.1736 | 0.1708 | 0.1333 | 0.1250 | 0.0986 | 0.0986 |
| SIMPLSR | 0.1792 | 0.1486 | 0.1306 | 0.1194 | 0.0833 | 0.0681 |
| SPLPLSR | 0.1861 | 0.1722 | 0.1264 | 0.1153 | 0.1028 | 0.0750 |
| PLSRGGR | 0.1000 | 0.0903 | 0.0986 | **0.0292** | **0.0292** | 0.0306 |
| DAPLSR | **0.0917** | **0.0667** | **0.0694** | **0.0292** | **0.0278** | **0.0278** |

To better analyze model performance, the proposed model and other comparative methods are evaluated on the COIL-20 dataset using several performance metrics. Table 4 clearly presents these comparative results, with the best performance highlighted in bold. In the DAPLSR model experiments, the original samples are augmented by 120%. A detailed comparison reveals that the data augmentation-based model exhibits higher accuracy and superior performance in object image recognition tasks, significantly outperforming other methods. This finding validates the effectiveness of data augmentation in enhancing model performance and provides strong support for future research.

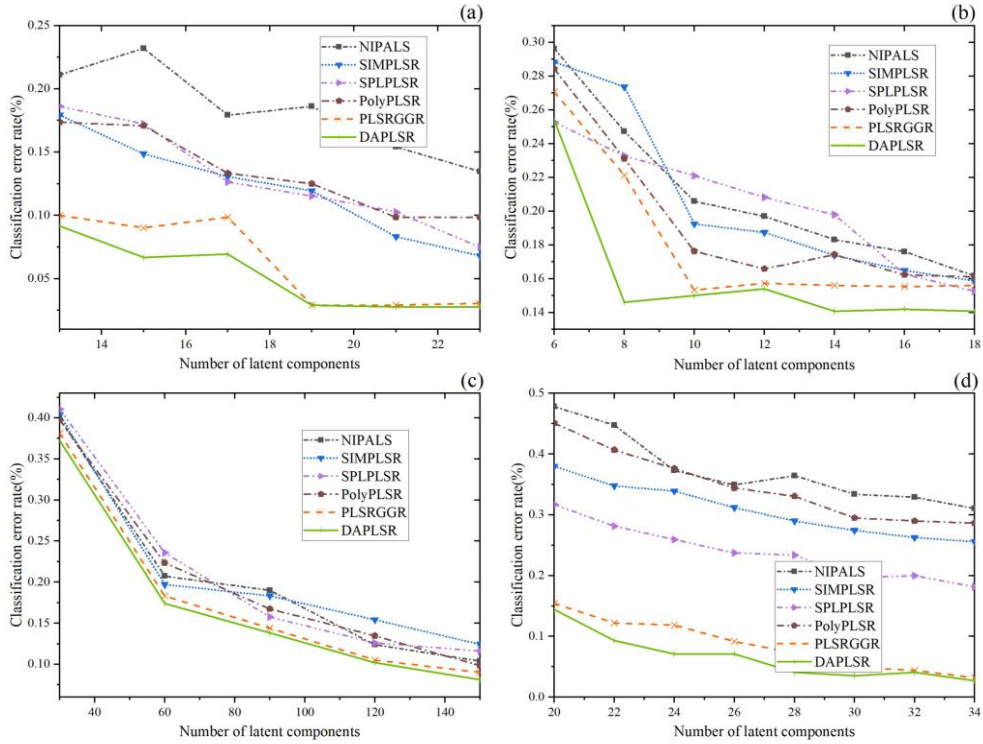

Figure 2. (a) COIL-20, (b) USPS, (c) Brodatz, and (d) EYaleB datasets show classification error rate line charts for different methods, where lower values indicate better model performance

Table 4. COIL-20 Dataset Performance Metrics

| Method | Accuracy | G-mean | Precision | Recall | F-measure |
|---|---|---|---|---|---|
| NIPALS | 0.8222 | 0.8838 | 0.8506 | 0.8222 | 0.8021 |
| PolyPLSR | 0.8847 | 0.9289 | 0.9003 | 0.8847 | 0.8736 |
| SIMPLSR | 0.8944 | 0.9342 | 0.9085 | 0.8944 | 0.8830 |
| SPLPLSR | 0.8958 | 0.9365 | 0.9070 | 0.8958 | 0.8857 |
| PLSRGGR | 0.9486 | 0.9716 | 0.9532 | 0.9486 | 0.9478 |
| DAPLSR | **0.9694** | **0.9833** | **0.9704** | **0.9694** | **0.9690** |

Additionally, experiments are conducted on the USPS handwritten digit dataset. During the DAPLSR model experiments, SMOTE is used to oversample the original samples, increasing the sample size by 40%. The categorization outcomes are displayed in Table 5.

Table 5. USPS Dataset Classification Error Rates

| c | 6 | 8 | 10 | 12 | 14 | 16 | 18 |
|---|---|---|---|---|---|---|---|
| NIPALS | 0.2962 | 0.2473 | 0.2058 | 0.1969 | 0.1832 | 0.1760 | 0.1617 |
| PolyPLSR | 0.2843 | 0.2311 | 0.1762 | 0.1658 | 0.1743 | 0.1623 | 0.1611 |
| SIMPLSR | 0.2883 | 0.2737 | 0.1925 | 0.1874 | 0.1739 | 0.1650 | 0.1587 |
| SPLPLSR | 0.2528 | 0.2328 | 0.2210 | 0.2082 | 0.1979 | 0.1628 | 0.1526 |
| PLSRGGR | 0.2707 | 0.2213 | 0.1533 | 0.1573 | 0.1560 | 0.1553 | 0.1560 |
| DAPLSR | **0.2180** | **0.1460** | **0.1393** | **0.1400** | **0.1407** | **0.1420** | **0.1407** |

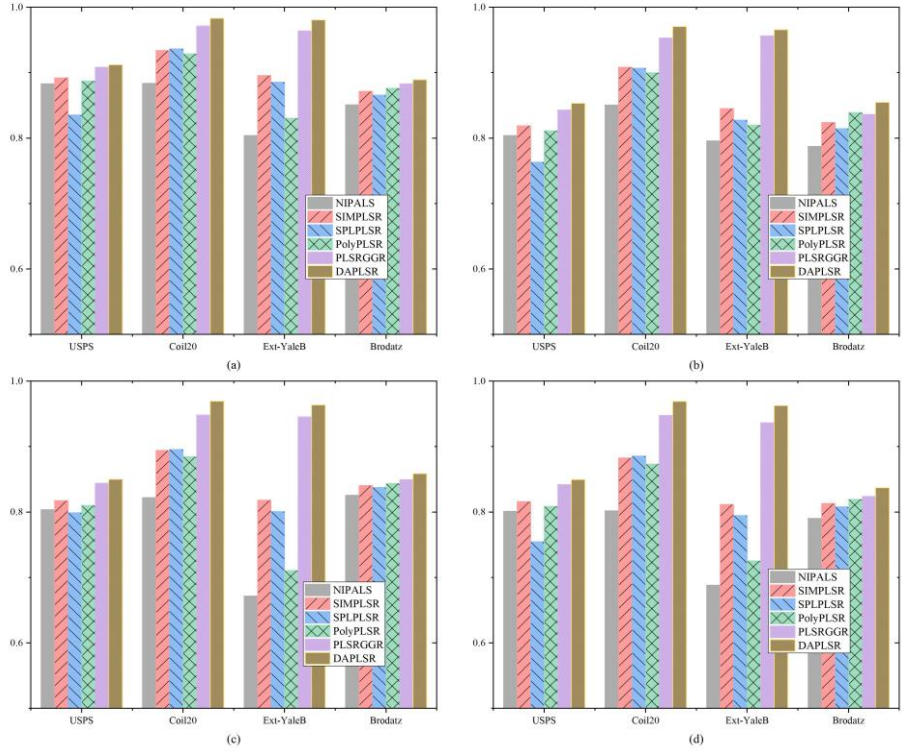

Figure 3. shows the results of different models on the USPS, COIL-20, EYaleB, and Brodatz datasets for (a) G-mean, (b) Precision, (c) Recall, and (d) F-measure. Higher values indicate better model performance

According to the experimental findings, the proposed DAPLSR model provides reduced classification error rates for the chosen component count. For instance, when retaining 12 components, the classification error rate is reduced by 1.73% compared to the second-best method. This demonstrates the value of data augmentation in predictive models. The classification performance is significantly improved when the PLSR model and data augmentation techniques are combined. This result not only emphasizes the critical role of data augmentation in enhancing model performance but also validates the model's generalization capability.

In addition, Table 6 presents other evaluation metrics for the comparative methods on the USPS dataset. The proposed model generally achieves the best results among the comparative methods, further confirming that the data augmentation-based partial least squares model exhibits superior classification performance in handwritten digit classification tasks.

Table 6.    USPS Dataset Performance Metrics

| Method | Accuracy | G-mean | Precision | Recall | F-measure |
|---|---|---|---|---|---|
| NIPALS | 0.8027 | 0.8833 | 0.8044 | 0.8037 | 0.8013 |
| PolyPLSR | 0.8120 | 0.8877 | 0.8115 | 0.8102 | 0.8090 |
| SIMPLSR | 0.8187 | 0.8922 | 0.8190 | 0.8177 | 0.8160 |
| SPLPLSR | 0.7908 | 0.8357 | 0.7634 | 0.7992 | 0.7547 |
| PLSRGGR | 0.8440 | 0.9085 | 0.8431 | 0.8444 | 0.8420 |
| DAPLSR | **0.8520** | **0.9120** | **0.8534** | **0.8499** | **0.8495** |

Table 7 presents the experimental results of the DAPLSR model and its comparative methods on the Brodatz texture dataset. The data indicate that the proposed DAPLSR model achieves superior classification performance compared to other models. It can be inferred that incorporating data augmentation techniques into the PLSR model enhances sample domain diversity and compensates for potential information loss in PLSR applications.

Table 7.    Brodatz Dataset Classification Error Rates

| c | 30 | 60 | 90 | 120 | 150 |
|---|----|----|----|-----|-----|
| NIPALS | 0.3986 | 0.2074 | 0.1901 | 0.1240 | 0.1040 |
| PolyPLSR | 0.3975 | 0.2235 | 0.1675 | 0.1347 | 0.0984 |
| SIMPLSR | 0.4038 | 0.1972 | 0.1833 | 0.1543 | 0.1243 |
| SPLPLSR | 0.4106 | 0.2358 | 0.1576 | 0.1261 | 0.1161 |
| PLSRGGR | 0.3814 | 0.1832 | 0.1441 | 0.1051 | 0.0901 |
| DAPLSR | **0.3724** | **0.1742** | **0.1381** | **0.1021** | **0.0811** |

Table 8 reveals that increasing the number of data samples improves classification accuracy and reduces error rates. This analysis highlights the importance of sample diversity in improving model performance on texture datasets. Particularly for complex datasets, increasing the number of samples can significantly enhance classification accuracy. The superior performance of the DAPLSR model on the Brodatz dataset further validates its effectiveness in texture classification tasks.

Table 8.    Brodatz Dataset Performance Metrics

| Method | Accuracy | G-mean | Precision | Recall | F-measure |
|--------|----------|--------|-----------|--------|-----------|
| NIPALS | 0.8258 | 0.8511 | 0.7876 | 0.8258 | 0.7906 |
| PolyPLSR | 0.8438 | 0.8763 | 0.8392 | 0.8438 | 0.8197 |
| SIMPLSR | 0.8408 | 0.8717 | 0.8241 | 0.8408 | 0.8132 |
| SPLPLSR | 0.8378 | 0.8660 | 0.8148 | 0.8378 | 0.8082 |
| PLSRGGR | 0.8499 | 0.8833 | 0.8365 | 0.8498 | 0.8241 |
| DAPLS | **0.8589** | **0.8893** | **0.8549** | **0.8588** | **0.8371** |

## 5. CONCLUSION

This paper proposes a novel data augmentation Partial Least Squares Regression model via manifold optimization. The DAPLSR model employs data augmentation to enrich the dataset, thereby enhancing the model's classification performance and generalization capability. Concurrently, the local geometric structure of the manifold is exploited, and a manifold optimization method is introduced to solve the proposed objective function, resulting in the optimal projection matrix. This process further refines all factors within the PLSR model, thereby improving its overall performance. The proposed DAPLSR model demonstrates superior classification performance and remarkable evaluation metrics on the EYaleB, COIL-20, USPS, and Brodatz texture datasets, significantly surpassing existing methods.

## ACKNOWLEDGMENTS

The work is supported by the National Natural Science Foundation of China under Grant No. 61906175 and No. 62172023, the Key Research Project of Colleges and Universities of Henan Province No. 22A520013, Henan Provincial Natural Science Foundation Youth Science Fund No. 242300420694 and No. 242300421220, the Science and Technology Project of Henan

## ABBREVIATIONS

The following abbreviations are used in this manuscript:

DAPLSR Data Augmentation Partial Least Squares Regression
SMOTE Synthetic Minority Over-sampling Technique
PLSR Partial Least Squares Regression
VDM Value Difference Metric
PLS Partial Least Squares
PCA Principal Component Analysis

## ACKNOWLEDGEMENTS

The authors would like to thank everyone, just everyone!

## REFERENCES

[1]     P. Sanskruti. "Marigold flower blooming stage detection in complex scene environment using faster RCNN with data augmentation." International Journal of Advanced Computer Science and Applications, vol. 14, no. 3, 2023.

[2]     M. Naseriparsa and M. M. Riahi Kashani. "Combination of PCA with SMOTE resampling to boost the prediction rate in lung cancer dataset." arXiv preprint arXiv:1403.1949, 2014.

[3]     M. A. Haque Farquad and I. Bose. "Preprocessing unbalanced data using support vector machine." Decision Support Systems, vol. 53, no. 1, pp. 226--233, 2012.

[4]     J. Tian, H. Gu, and W. Liu. "Imbalanced classification using support vector machine ensemble." Neural Computing and Applications, vol. 20, pp. 203--209, 2011.

[5]     J. Xie and Z. Qiu. "The effect of imbalanced data sets on LDA: A theoretical and empirical analysis." Pattern Recognition, vol. 40, no. 2, pp. 557--562, 2007.

[6]     W. Shao, X. Tian, P. Wang, X. Deng, and S. Chen. "Online soft sensor design using local partial least squares models with adaptive process state partition." Chemometrics and Intelligent Laboratory Systems, vol. 144, pp. 108--117, 2015.

[7]     S. Wold, M. Sj"ostr"om, and L. Eriksson. "PLS-regression: A basic tool of chemometrics." Chemometrics and Intelligent Laboratory Systems, vol. 58, no. 2, pp. 109--130, 2001.

[8]     N. A. M. Salleh and M. S. Hassan. "Discrimination of lard and other edible fats after heating treatments using Partial Least Square Regression (PLSR), Principal Component Regression (PCR) and Linear Support Vector Machine Regression (SVMR)." Journal of Physics: Conference Series, 2019.

[9]     J. Huang, X. Li, and G. Wang. "Maximum principles for a class of partial information risk-sensitive optimal controls." IEEE Transactions on Automatic Control, vol. 55, no. 6, pp. 1438--1443, 2010.

[10]    J. Marques and E. Dam. "Texture analysis by a PLS based method for combined feature extraction and selection." In Proceedings of the Second International Workshop on Machine Learning in Medical Imaging, MICCAI 2011, Toronto, Canada, September 18, pp. 109--116, 2011.

[11]    X. Gan, J. Duanmu, J. Wang, and W. Cong. "Anomaly intrusion detection based on PLS feature extraction and core vector machine." Knowledge-Based Systems, vol. 40, pp. 1--6, 2013.

[12]    H. Han, W. Wang, and B. Mao. "Borderline-SMOTE: A new over-sampling method in imbalanced data sets learning." In Proceedings of the International Conference on Intelligent Computing, pp. 878--887, 2005.

[13]    R. Blagus and L. Lusa. "SMOTE for high-dimensional class-imbalanced data." BMC Bioinformatics, vol. 14, pp. 1--16, 2013.

[14] T. R. Payne and P. Edwards. "Implicit feature selection with the value difference metric." In Proceedings of the European Conference on Artificial Intelligence, ECAI-98, 1998.

[15] T. Sugimura, E. Arnold, S. English, and J. Moore. "Chronic suprapubic catheterization in the management of patients with spinal cord injuries: Analysis of upper and lower urinary tract complications." BJU International, vol. 101, no. 11, pp. 1396--1400, 2008.

[16] F. Camarrone and M. M. Van Hulle. "Fast multiway partial least squares regression." IEEE Transactions on Biomedical Engineering, vol. 66, no. 2, pp. 433--443, 2018.

[17] R. Rosipal and N. Kr{"a}mer. "Overview and recent advances in partial least squares." In Proceedings of the International Statistical and Optimization Perspectives Workshop "Subspace, Latent Structure and Feature Selection", pp. 34--51, 2005.

[18] A. M. Martinez and A. C. Kak. "PCA versus LDA." IEEE Transactions on Pattern Analysis and Machine Intelligence, vol. 23, no. 2, pp. 228--233, 2001.

[19] K. Raslan, A. Alsharkawy, and K. Raslan. "HHO-SMOTe: Efficient sampling rate for Synthetic Minority Oversampling technique based on Harris Hawk optimization." International Journal of Advanced Computer Science and Applications, vol. 14, no. 10, 2023.

[20] E. Lashgari, D. Liang, and U. Maoz. "Data augmentation for deep-learning-based electroencephalography." Journal of Neuroscience Methods, vol. 346, pp. 108885, 2020.

[21] S. De Jong. "SIMPLS: An alternative approach to partial least squares regression." Chemometrics and Intelligent Laboratory Systems, vol. 18, no. 3, pp. 251--263, 1993.

[22] P. Absil, R. Mahony, and R. Sepulchre. Optimization algorithms on matrix manifolds. Princeton University Press, 2008.

[23] Z. Gao, Y. Wu, X. Fan, M. Harandi, and Y. Jia. "Learning to optimize on riemannian manifolds." IEEE Transactions on Pattern Analysis and Machine Intelligence, vol. 45, no. 5, pp. 5935--5952, 2022.

[24] A. Douik, X. Liu, T. Ballal, T. Y. Al-Naffouri, and B. Hassibi. "Precise 3-D GNSS attitude determination based on riemannian manifold optimization algorithms." IEEE Transactions on Signal Processing, vol. 68, pp. 284--299, 2019.

[25] S. Boyd, N. Parikh, E. Chu, B. Peleato, J. Eckstein, and others. "Distributed optimization and statistical learning via the alternating direction method of multipliers." Foundations and Trends in Machine Learning, vol. 3, no. 1, pp. 1--122, 2011.

[26] R. Rosipal. "Nonlinear partial least squares: An overview." Chemoinformatics and Advanced Machine Learning Perspectives: Complex Computational Methods and Collaborative Techniques, pp. 169--189, 2011.

[27] R. M. Balabin and E. I. Lomakina. "Support vector machine regression (SVR/LS-SVM)—An alternative to neural networks (ANN) for analytical chemistry? Comparison of nonlinear methods on near infrared (NIR) spectroscopy data." Analyst, vol. 136, no. 8, pp. 1703--1712, 2011.

[28] S. Wold. "Nonlinear partial least squares modelling II. Spline inner relation." Chemometrics and Intelligent Laboratory Systems, vol. 14, no. 1-3, pp. 71--84, 1992.

[29] H. Chen, Y. Sun, J. Gao, Y. Hu, and B. Yin. "Solving partial least squares regression via manifold optimization approaches." IEEE Transactions on Neural Networks and Learning Systems, vol. 30, no. 2, pp. 588--600, 2018.

[30] L. Qiao, S. Chen, and X. Tan. "Sparsity preserving projections with applications to face recognition." Pattern Recognition, vol. 43, no. 1, pp. 331--341, 2010.

[31] S. A. Nene, S. K. Nayar, and H. Murase. "Columbia object image library (coil-100)." Columbia University, vol. 62, 1996.

[32] J. J. Hull, "A database for handwritten text recognition research," IEEE Transactions on Pattern Analysis and Machine Intelligence, vol. 16, no. 5, pp. 550--554, 1994.

[33] X. Qi, R. Xiao, C. Li, Y. Qiao, J. Guo, and X. Tang, "Pairwise rotation invariant co-occurrence local binary pattern," IEEE Transactions on Pattern Analysis and Machine Intelligence, vol. 36, no. 11, pp. 2199--2213, 2014.

## AUTHORS

**Haoran Chen** is a lecturer at Zhengzhou Institute of Light Industry. He graduated from the Department of Mathematics of Zhengzhou University with his master's degree in 2010 and from Beijing University of Technology with his PhD in 2019. His main research areas involve artificial intelligence, machine learning, pattern recognition, and manifold optimization.

**Jiapeng Liu** received her BS degree from Zhengzhou University of Light Industry in 2022. He is currently pursuing his MS degree at Zhengzhou University of Light Industry. His current research interests include multimodal sentiment analysis, pattern recognition and image processing.

**Jiafan Wang** received her MS degree from Zhengzhou University of Light Industry in 2024. Her current research interests include pattern recognition and image processing.

**Wenjun Shi** received B.Sc. degree from Henan Agricultural University in 2011, master degree and doctoral degree from the Tiangong University in 2014 and 2021. Now she is a lecturer at Zhengzhou University of Light Industry. Her main research interests include parallel and distributed computing, edge intelligence and data science.